\documentclass{article}

\usepackage{PRIMEarxiv}

\usepackage[utf8]{inputenc} % allow utf-8 input
\usepackage[T1]{fontenc}    % use 8-bit T1 fonts
\usepackage{hyperref}       % hyperlinks
\usepackage{url}            % simple URL typesetting
\usepackage{booktabs}       % professional-quality tables
\usepackage{amsfonts}       % blackboard math symbols
\usepackage{nicefrac}       % compact symbols for 1/2, etc.
\usepackage{microtype}      % microtypography
\usepackage{lipsum}
\usepackage{fancyhdr}       % header
\usepackage{graphicx}       % graphics
\graphicspath{{media/}}     % organize your images and other figures under media/ folder

\title{Mind meets machine: Unravelling GPT-4's cognitive psychology
}

\author{
  Sifatkaur Dhingra\\
  Department of Psychology, Nowrosjee Wadia College \\
  Pune, India\\
  \texttt{sifatkaurd13@gmail.com} \\
   \And
  Manmeet Singh \\
  Indian Institute of Tropical Meteorology \\
  Pune, India\\
  \texttt{manmeet.cat@tropmet.res.in} \\
  \And
  Vaisakh SB \\
  Indian Institute of Tropical Meteorology \\
  Pune, India\\
  \texttt{vaisakh.sb@tropmet.res.in} \\
  \And
  Neetiraj Malviya \\
  Defence Institute Of Advanced Technology \\
  Pune, India\\
  \texttt{neetirajmalviya@gmail.com} \\
  \And
  Sukhpal Singh Gill \\
  Queen Mary University of London \\
  London, United Kingdom\\
  \texttt{s.s.gill@qmul.ac.uk} \\
}

\begin{document}
\maketitle

\begin{abstract}
Cognitive psychology delves on understanding perception, attention, memory, language, problem-solving, decision-making, and reasoning. Large language models (LLMs) are emerging as potent tools increasingly capable of performing human-level tasks. The recent development in the form of GPT-4 and its demonstrated success in tasks complex to humans exam and complex problems has led to an increased confidence in the LLMs to become perfect instruments of intelligence. Although GPT-4 report has shown performance on some cognitive psychology tasks, a comprehensive assessment of GPT-4, via the existing well-established datasets is required. In this study, we focus on the evaluation of GPT-4's performance on a set of cognitive psychology datasets such as CommonsenseQA, SuperGLUE, MATH and HANS. In doing so, we understand how GPT-4 processes and integrates cognitive psychology with contextual information, providing insight into the underlying cognitive processes that enable its ability to generate the responses. We show that GPT-4 exhibits a high level of accuracy in cognitive psychology tasks relative to the prior state-of-the-art models. Our results strengthen the already available assessments and confidence on GPT-4's cognitive psychology abilities. It has significant potential to revolutionize the field of AI, by enabling machines to bridge the gap between human and machine reasoning. 
\end{abstract}

% keywords can be removed
% \keywords{Urban downscaling, Deep learning, Smart city}

\section{Introduction}
\label{sec:intro}
Cognitive psychology aims to decipher how humans learn new things, retain knowledge, and recall it when needed. Cognitive psychologists seek to understand how the mind works by conducting studies on people's thoughts and actions and by using other experimental methods like brain imaging and computer modelling. Understanding the human mind and developing our cognitive skills to excel in a variety of areas is the ultimate objective of cognitive psychology. 
Language models have come a long way since the first statistical models for modelling language were introduced. With the advent of deep learning and the availability of large amounts of data, recent years have seen a rapid evolution of language models that have achieved human-like performance on many language tasks. Large Language Models (LLMs) are a type of artificial intelligence framework that have garnered significant attention in recent years due to their remarkable language processing capabilities (Harrer 2023). These models are trained on vast amounts of text data and are able to generate coherent, human-like responses to natural language queries. One of the key features of LLMs is their ability to generate novel and creative responses to text-based prompts, which has led to their increasing use in fields such as chatbots, question answering systems, and language translation. The use of self-attention has been a key factor in this success, as it allows for more efficient and accurate modeling of long-range dependencies within the input sequence, resulting in better performance compared to traditional RNN-based models. LLMs have demonstrated impressive performance on a wide range of language tasks, including language modeling, machine translation, sentiment analysis, and text classification. These capabilities have led to the increased use of LLMs in various fields, including language-based customer service, virtual assistants, and creative writing. 
One of the key areas measuring intelligence in humans, other species and machines is the cognitive psychology. There are several tasks that are considered to be the benchmarks for testing cognitive psychology. Some of them are text interpretation, computer vision, planning and reasoning. For cognitive psychology to work, we rely on a complex and potent social practise: the attribution and assessment of thoughts and actions  
\cite{madaan2022text}. The scientific psychology of cognition and behaviour, a relatively recent innovation, focuses primarily on the information-processing mechanisms and activities that characterise human cognitive and behavioural capabilities. Researchers have attempted to create systems that could use natural language to reason about their surroundings \cite{mccarthy1959basis} or that could use a world model to get a more profound comprehension of spoken language \cite{winograd1972understanding}.
The report introducing GPT-4 \cite{gpt4openai} has tested the HellaSwag \cite{zellers2019hellaswag} and WinoGrande \cite{sakaguchi2021winogrande} datasets for cognitive psychology. Although, these tests are relevant, they lack the sophistication required to understand deep heuristics of GPT-4. Hellaswag entails the task of finishing a sentence and WinoGrande involves identifying the correct noun for the pronouns in a sentence, which are quite simple. Other tasks and standardized datasets \cite{li2022systematic} which test the psychology are needed in order to perform a comprehensive assessment of cognitive psychology for GPT-4. Moreover GPT-4 needs to go through complex reasoning tasks than just predicting the last word of the sentence such as in Hellaswag, to emerge as a model capable of high-level intelligence. \cite{shiffrin2023probing} note that SuperGLUE \cite{wang2019superglue}, CommonsenseQA \cite{talmor2018commonsenseqa}, MATH \cite{hendrycks2021measuring} and HANS \cite{mccoy2019right} are four such datasets that are needed to be tested for a comprehensive cognitive psychology evaluation of AI models. In this study, we evaluate the performance of GPT-4 on the SuperGLUE, CommonsenseQA, MATH and HANS datasets. This is a work in progress and we are performing continuous tests with the other datasets as suggested by \cite{shiffrin2023probing}. Our study can be used to build up higher-order psychological tests using GPT-4. 

\section{Datasets and Methodology}
In this study, four datasets have been used to test the cognitive psychology capabilities of GPT-4. The four datasets are CommonsenseQA, MATH, SuperGLUE and HANS. They are described as below:

\subsection{CommonsenseQA}
CommonsenseQA is a dataset composed for testing commonsense reasoning. There are 12,247 questions in the dataset, each with 5 possible answers. Workers using Amazon's Mechanical Turk were used to build the dataset. The goal of the dataset is to evaluate the commonsense knowledge using CONCEPTNET to generate difficult questions. The language model tested in the CommonsenseQA paper has an accuracy of 55.9 \% whereas the authors report that human accuracy on the dataset is around 89 \%.

\subsection{MATH}
The MATH dataset includes almost 12,500 problems from scholastic mathematics contests. Machine learning models take a mathematical problem as input and produce an answer-encoding sequence, such as $frac23$. After normalisation, their answers are distinct, therefore MATH may be evaluated using exact match instead of heuristic metrics like BLEU. Problems in seven different areas of mathematics, including geometry, are categorised by complexity from 1 to 5, and diagrams can be expressed in text using the Asymptote language. This allows for a nuanced evaluation of problem-solving skills in mathematics across a wide range of rigour and content. Problems now have comprehensive, detailed, step-by-step answers. To improve learning and make model outputs more interpretable, models can be trained on these to develop their own step-by-step solutions. The MATH dataset presents a significant challenge, with accuracy rates for big language models ranging from 3.0\% to 6.9\%. Models attain up to 15\% accuracy on the least difficulty level and can develop step-by-step answers that are coherent and on-topic even when erroneous, suggesting that they do possess some mathematical knowledge despite their low accuracies. The results of human evaluations on MATH show that it may be difficult for humans as well; a computer science PhD student who does not really like mathematics scored about 40\%, while a three-time IMO gold medallist scored 90\%.
\subsection{SuperGLUE}
SuperGLUE is an updated version of the GLUE benchmark that includes a more challenging set of language understanding tasks. Using the gap between human and machine performance as a metric, SuperGLUE improves upon the GLUE benchmark by defining a new set of difficult Natural Language Understanding (NLU) problems. About half of the tasks in the SuperGLUE benchmark have fewer than 1k instances, and all but one have fewer than 10k examples, highlighting the importance of different task formats and low-data training data problems. As compared to humans, SuperGLUE scores roughly 20 points worse when using BERT as a baseline in the original study. To get closer to human-level performance on the benchmark, the authors argue that advances in multi-task, transfer, and unsupervised/self-supervised learning approaches are essential.
\subsection{HANS}
The strength of neural networks lies in their ability to analyse a training set for statistical patterns and then apply those patterns to test instances that come from the same distribution. This advantage is not without its drawbacks, however, as statistical learners, such as traditional neural network designs, tend to rely on simplistic approaches that work for the vast majority of training samples rather than capturing the underlying generalisations. The loss function may not motivate the model to learn to generalise to increasingly difficult scenarios in the same way a person would if heuristics tend to produce mostly correct results. This problem has been observed in several applications of AI. Contextual heuristics mislead object-recognition neural networks in computer vision, for example; a network that can accurately identify monkeys in a normal situation may mistake a monkey carrying a guitar for a person, since guitars tend to co-occur with people but not monkeys in the training set. Visual question answering systems are prone to the same heuristics. This problem is tackled by HANS (Heuristic Analysis for NLI Systems), which uses heuristics to determine if a premise sentence entails (i.e., suggests the truth of) a hypothesis sentence. Neural Natural Language Inference (NLI) models have been demonstrated to learn shallow heuristics based on the presence of specific words, as has been the case in other fields. As not often appears in the instances of contradiction in normal NLI training sets, a model can categorise all inputs containing the word not as contradiction. HANS prioritises heuristics that are founded on elementary syntactic characteristics. Think about the entailment-focused phrase pair below:

Premise: The judge was paid by the actor.

Hypothesis: The actor paid the judge.

An NLI system may accurately label this example not by deducing the meanings of these lines but by assuming that the premise involves any hypothesis whose terms all occur in the premise. Importantly, if the model is employing this heuristic, it will incorrectly classify the following as entailed even when it is not.

Premise: The actor was paid by the judge.

Hypothesis: The actor paid the judge.

HANS is intended to detect the presence of such faulty structural heuristics. The authors focus on the lexical overlap, subsequence, and component heuristics. These heuristics are not legitimate inference procedures despite often producing correct labels. Rather than just having reduced overall accuracy, HANS is meant to ensure that models using these heuristics fail on specific subsets of the dataset. Four well-known NLI models, including BERT, are compared and contrasted using the HANS dataset. For this dataset, all models significantly underperformed the chance distribution, with accuracy just exceeding 0\% in most situations. 
\subsection{Methodology}
We test the four datasets as described above to test the cognitive psychology capabilities of GPT-4. The model is accessed using the ChatGPT-Plus offered by OpenAI. We evaluate these models as shown in the results and discussion section.

\begin{figure*}
  \centering
  \includegraphics[width=0.95\linewidth]{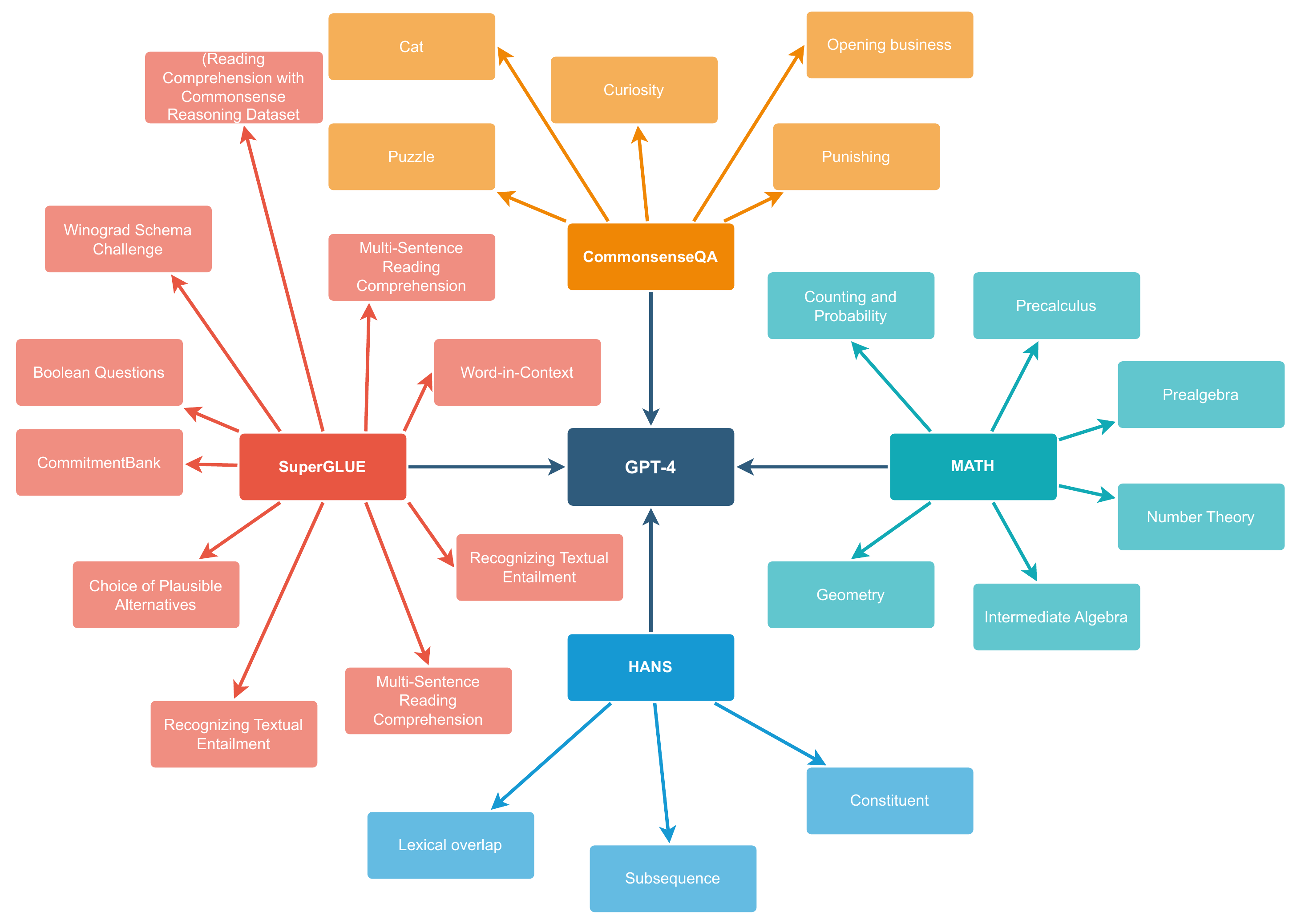}

  \caption{Datasets used in the study with the different categories contained in them.}
  \label{fig-1}
\end{figure*}

\begin{figure*}
  \centering
  \includegraphics[width=0.95\linewidth]{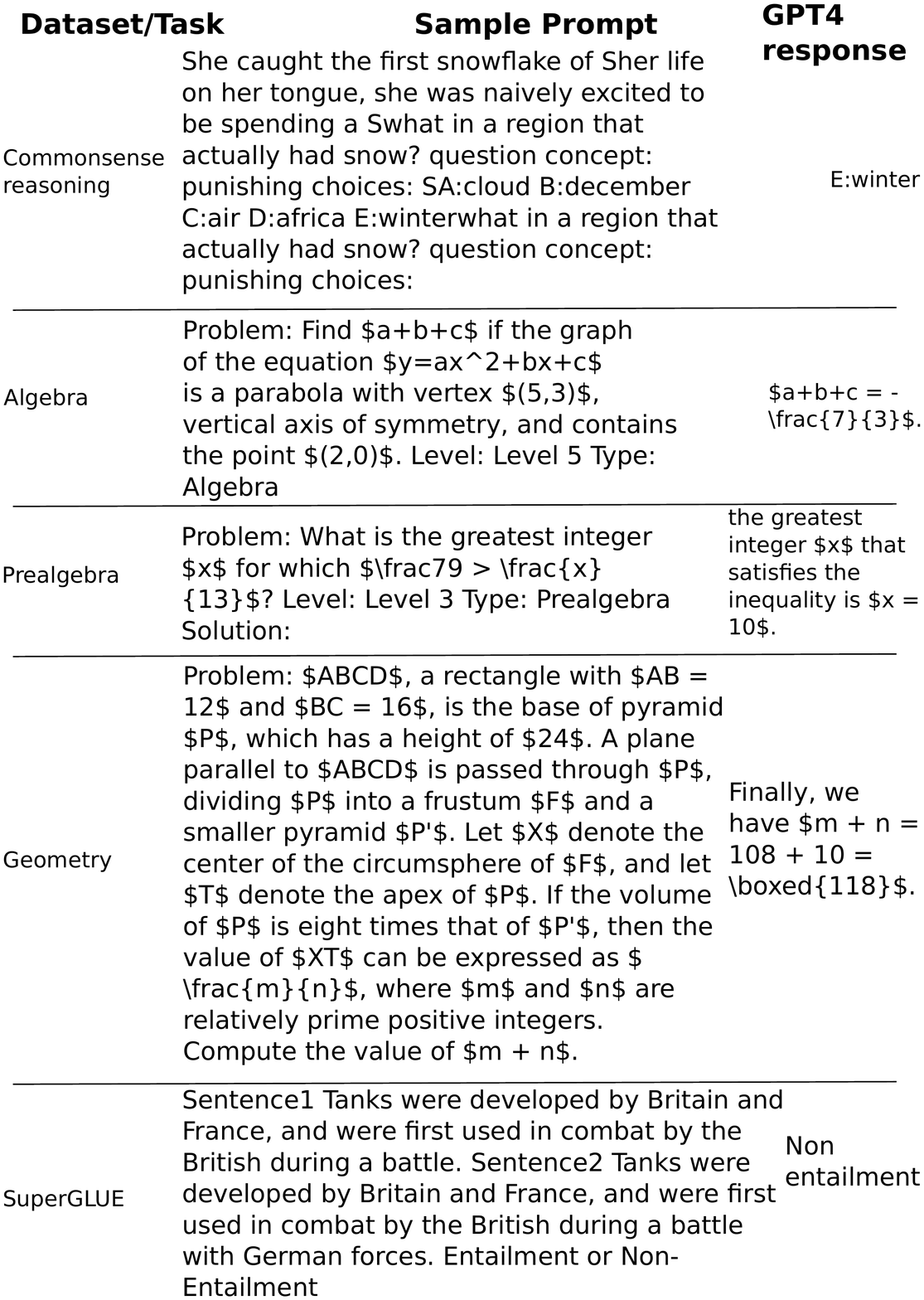}

  \caption{Examples of sample prompts and the respective responses of GPT4 on CommonsenseQA, MATH and SuperGLUE datasets}
  \label{fig-2}
\end{figure*}
%\section{Methodology}

\section{Results}
We will first discuss the human and machine skill of the different models traditionally used in the datasets used to test cognitive psychology. As compared to humans, SuperGLUE scores roughly 20 points worse when using BERT as a baseline in the original study. To get closer to human-level performance on the benchmark, the authors argue that advances in multi-task, transfer, and unsupervised/self-supervised learning approaches are essential. The language model tested in the CommonsenseQA paper has an accuracy of 55.9 \% whereas the authors report that human accuracy on the dataset is around 89 \%. The accuracy of humans on HANS dataset ranged from 76-97 \% and the authors show that the BERT model performed below 10 \% on the non-entailment category. The human performance on MATH varied from 40-90 \% and GPT-2/GPT-3 showed accuracies below 10 \%.

Figure 1 shows that GPT-4 has an accuracy of 83.2 \% on CommonSenseQA,  data, we find that GPT-4 has an accuracy of around 84\%, 82 \% on prealgebra, 35\% on geometry,  100\% on HANS and 91.2 \% on SuperGLUE. It is to be noted that the perfect results on HANS data might be because all the examples used are of non-entailment, as the model might be memorizing this particular heuristic. The experiments to generate GPT-4 results with mixed data from HANS are ongoing.

\begin{figure*}
  \centering
  \includegraphics[width=0.6\linewidth]{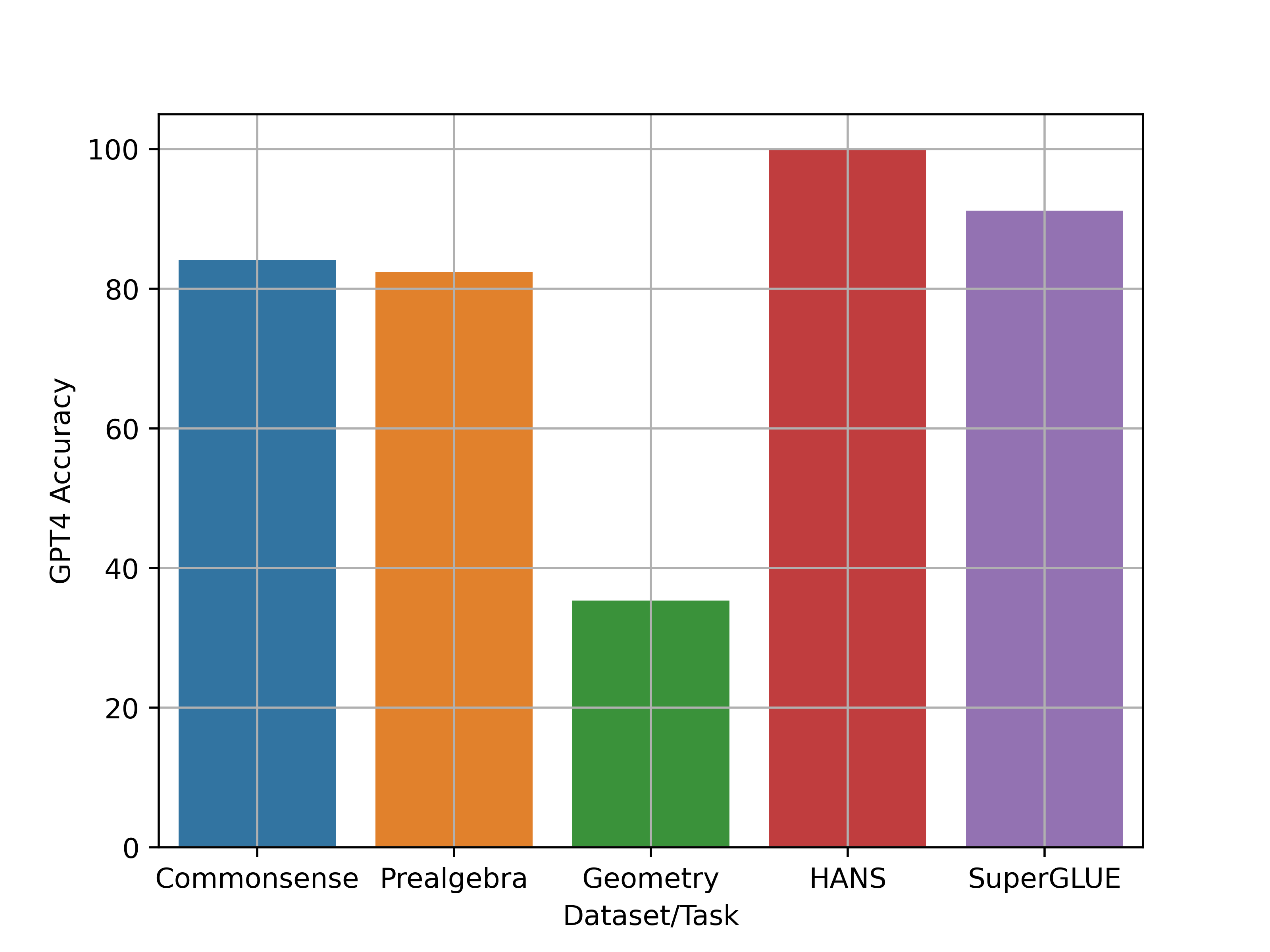}

  \caption{Accuracy of GPT4 on cognitive psychology tasks}
  \label{fig-3}
\end{figure*}
\section{Conclusions}
GPT-4, which is a state-of-the-art large language model, is a revolution in the field of psychology since it gives psychologists unprecedented resources to use in their studies and work. This sophisticated AI model offers psychologists and psychiatrists to learn more about the human mind and come up with novel treatment theories and approaches. It provides an avenue for improved efficacy of psychological therapies and allowing professionals to spend more time with clients, leading to deeper and more fruitful therapeutic bonds. The potential applications of GPT-4 can only be realized if the model is thoroughly tested on basic tests of reasoning and cognition. Cognitive psychology enables the humans to perform various activities \cite{aher2022using} in their personal and professional lives. We show that the performance of GPT-4 greatly surpasses the language model used in the original studies from where the different datasets are sourced, thus it can make a tool of day-to-day utility for psychologists. This development can lead to cascading benefits in addressing the mental health challenges faced by today's society.

\bibliographystyle{unsrt}  
\bibliography{templatePRIME}

\end{document}